\documentclass[10pt, a4paper]{article}

\usepackage[final]{lrec-coling2024} 
\usepackage{amsmath}
\usepackage{fancyvrb}
\usepackage{amssymb}
\usepackage{booktabs}
\usepackage{listings}
\usepackage{multirow}

\usepackage{algorithm}
\usepackage[noend]{algpseudocode}
\usepackage{enumitem}
\usepackage{placeins}
\usepackage{fvextra}

\title{An LLM-Enhanced Adversarial Editing System \\for Lexical Simplification}

\name{
Keren Tan\textsuperscript{1}, 
Kangyang Luo\textsuperscript{1}, 
Yunshi Lan\textsuperscript{1}\textsuperscript{*}\thanks{* Corresponding author.}, 
Zheng Yuan\textsuperscript{2}, 
Jinlong Shu\textsuperscript{1}
} 

\address{
\textsuperscript{1}School of Data Science \& Engineering, East China Normal University, Shanghai, China \\
\textsuperscript{2}Department of Informatics, King’s College London, U.K.\\
\{tankeren1020, 52205901003\}@stu.ecnu.edu.cn, yslan@dase.ecnu.edu.cn, \\
zheng.yuan@kcl.ac.uk, jlshu@admin.ecnu.edu.cn\\
}

\abstract{
Lexical Simplification (LS) aims to simplify text at the lexical level. 
Existing methods rely heavily on annotated data, making it challenging to apply in low-resource scenarios. 
In this paper, we propose a novel LS method without parallel corpora. 
This method employs an Adversarial Editing System with guidance from a confusion loss and an invariance loss to predict lexical edits in the original sentences.
Meanwhile, we introduce an innovative LLM-enhanced loss to enable the distillation of knowledge from Large Language Models (LLMs) into a small-size LS system. 
From that, complex words within sentences are masked and a Difficulty-aware Filling module is crafted to replace masked positions with simpler words.
At last, extensive experimental results and analyses on three benchmark LS datasets demonstrate the effectiveness of our proposed method.
 \\ \newline 
}

\begin{document}

\maketitleabstract

\section{Introduction}
\label{sec:intro}

Text Simplification (TS) is the process of simplifying a sentence while retaining its semantics as much as possible.
It enables to lower the difficulty level of the entire sentence and helps people with cognitive disabilities to understand by making the text more readable~\cite{paetzold2017survey}.
As a special category of TS tasks, Lexical Simplification (LS) restricts the simplification at the lexical level via replacing complex words with alternative simpler words, thus minimising the revision to the original sentences.

Conventional LS tasks broadly consist of two sub-tasks, namely Complex Word Identification (CWI) and Substitute Generation (SG), which focus on detecting complex words and generating alternative words, respectively.
So far, a panoply of efforts work on addressing the said LS tasks.
For example, early LS systems leverage a set of rules for identifying and substituting complex words with frequent synonyms from external databases (e.g., WordNet \cite{miller1990introduction}, but these methods suffer from limited flexibility and adaptability~\cite{kajiwara2013selecting}.
Recently, popular LS systems first train a model to detect the complex words in a sentence, and then use another model to predict the alternative words, collaborating together to eventually produce a simplified sentence~\cite{qiang2021lsbert, seneviratne2022cils, wilkens2022cental}.
However, the mentioned two-stage approaches have a heavy reliance on the annotation of CWI and SG sub-tasks, thereby impairing their applications.
In real-world scenarios, acquiring parallel corpora for LS tasks is a costly endeavor, not to mention the annotation of CWI and SG.
As such, we aim to develop an LS system without parallel corpora in this work, but we are also confronted with the following challenges:
(1) In the absence of annotated data, the above-mentioned supervised training approaches are inapplicable, making it considerably challenging to ensure the accuracy of simplification.
(2) Constructing the previous two-stage system for LS tasks without parallel corpora is problematic, as in such scenarios, models struggle to learn the transformation from complex sentences to simplified ones.

\begin{figure}[!t]
\centering
\includegraphics[width=0.48\textwidth]{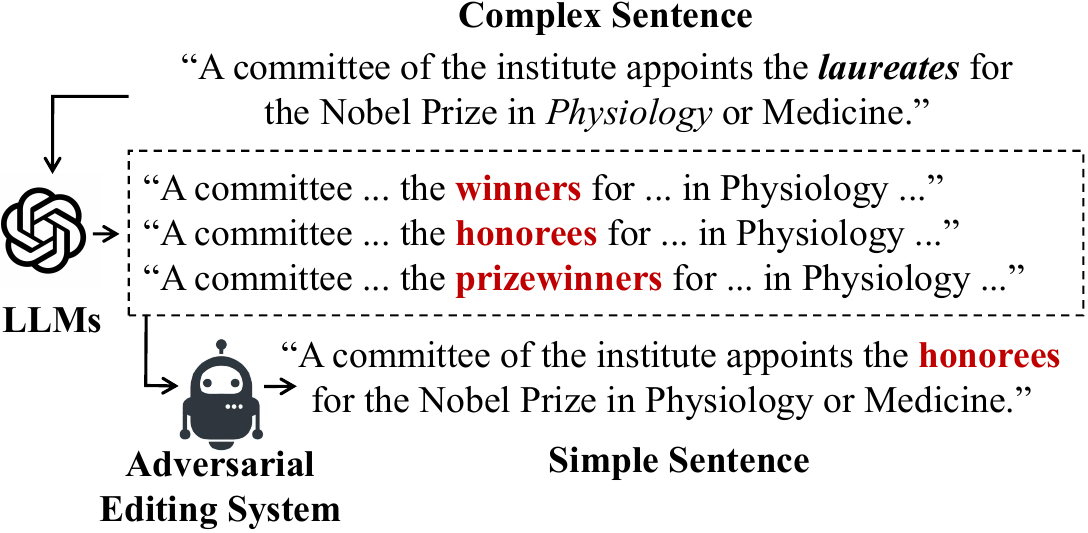} 
\vspace*{-3ex}
\caption{
The motivation of our LLM-enhanced Adversarial Editing System, that is, distilling the knowledge from LLMs to our small-size Adversarial Editing System.
In the complex sentence, correct complex words are in italic bold fonts, and red bold signifies differences between generated and original sentences.
}
\vspace*{-3ex}
\label{fig:motivation}
\end{figure}

Inspired by existing study in style transfer~\cite{wang2022text},
we develop an Adversarial Editing System to conduct lexical edits to the original sentence with the help of non-parallel corpora, where complex words are masked by the editing system, and the substitutions are generated via a cloze model following the two-stage approaches.
Nonetheless, striking a balance between semantic preservation and simplification degree remains a challenging endeavor. 
For example, in Figure~\ref{fig:motivation}, non-stylistic words \textit{``Physiology''} can be accidentally masked and it is hard to distinguish if \textit{``laureates''} is a complex word.

To this end, as the first attempt to LS without parallel corpora, we bring forward a new LS method 
dubbed as \textbf{LAE-LS} (\textbf{\underline{L}}LM-Enhanced \textbf{\underline{A}}dversarial \textbf{\underline{E}}diting System for \textbf{\underline{L}}exical \textbf{\underline{S}}implification), which involves two modules: Adversarial Editing and Difficulty-aware Filling.
Concretely, we employ an Adversarial Editing module to train an Edit Predictor for predicting lexical edits, which can be used to identify complex words within sentences.
To balance the preservation of semantics and the level of simplification, we introduce a confusion loss and an invariance loss to confuse the discriminator and preserve the semantics of the original sentence, respectively.
Particularly, an LLM-enhanced loss is tailored to extract supervision signals from Large Language Models (LLMs) (e.g., ChatGPT)~\cite{brown2020language, zhang2022opt,touvron2023llama}.
To be specific, we intricately design instructions to guide LLMs in identifying complex words within sentences, rather than directly commanding them to rewrite the original sentences, which can bypass changes to the sentence syntax. 
With this loss, high-quality knowledge from LLMs can be distilled into our Edit Predictor.
To avoid the overwhelming domination of the LLM-enhanced loss, we combine the above losses with weights when training.
Eventually, the Difficulty-aware Filling module is crafted to fill in the masked positions with alternative simple words.

In summary, we highlight our contributions as follows:
\begin{itemize}
\item 
We propose a novel LS method LAE-LS that is capable of making lexical edits to the original sentences without parallel corpora, thus rendering it feasible to perform LS tasks in low-resource scenarios.

\item 
To achieving tradeoff between the simplification degree and semantic preservation, we include confusion loss and invariance loss.
Furthermore, the LLM-enhanced loss is devised, enabling the distillation from LLMs to a small-size LS system.

\item Our method achieves SOTA results on three well-known LS datasets and yields competitive results to \texttt{GPT-3.5-turbo} even with a significantly smaller parameter size.

\end{itemize}

\section{Related Work}
\label{sec:related}

\noindent \textbf{Lexical Simplification}. 
The early methods for the LS task either resorted to threshold-based strategies~\cite{keskisarkka2012automatic} or relied on dictionaries to identify complex words within sentences~\cite{kajiwara2013selecting}, and took advantage of external resources (e.g., synonym dictionaries \citep{lesk1986automatic}, WordNet \cite{devlin1998use}, word embeddings \cite{mikolov2013efficient}) to provide substitutions for complex words.
However, these methods suffer from limited flexibility and adaptability \cite{paetzold2017survey}.
Shortly thereafter, a panoply of modifications for CWI and SG respectively have been proposed to alleviate said issues.
For CWI, \citet{gooding2018camb} shifts it towards training classifiers using supervised training with significant demand for feature engineering.
Recently, some works have achieved significant success by treating the CWI task as a sequence labeling task \cite{qiang2021lsbert, gooding2019recursive}.
For SG, the recent methods leverage the contextual comprehension capabilities of pre-trained models to generate substitutions for complex words.
For example, LSBert \citep{qiang2021lsbert} predicts alternative words for complex words by the BERT \cite{devlin2019bert} model.
SimpleBART \citep{sun2023teaching} augments pre-trained models through fine-tuning, enabling the effectively prediction for simpler words. 
ParaLS \cite{qiang-etal-2023-parals} fine-tunes a paraphraser to generate substitute candidates for complex words using two novel decoding strategies.
The said methods approach remarkably performance for the LS task, but they rely heavily on supervised training or a substantial amount of external linguistic resources.
On this account, our study is inspired by the aforementioned pitfalls.

\noindent \textbf{Adversarial Network and LLM}.
With the introduction of Generative Adversarial Networks (GAN) \citep{goodfellow2020generative}, adversarial learning has become ubiquitous in unsupervised training \citep{shen2017style, wang2022text}.
For example, KiS \citep{laban2021keep} extends Seq2Seq models to unparallel corpora scenarios where the decoder serves as the generator, and a component, which assesses whether a sentence is simplified or not, acts as the discriminator.
However, this method cannot guarantee that the syntax of the original sentence remains unchanged.
Accordingly, given the outstanding performance of LLMs \cite{zhang2022opt, brown2020language} in various linguistic tasks \cite{liang2023prompting,lan2023improving}, a stream of efforts has been explored the use of LLMs in TS, yielding promising results \citep{feng2023sentence, chi2023learning, sun2023teaching}.
Nevertheless, due to high resource consumption, time-intensive inference and over-simplification results, applying LLMs directly to LS is not practical.
Yet, we argue that using LLMs as an enhancement tool or component remains a viable option.
To the best of our knowledge, how to distil knowledge from LLMs for augmenting LS tasks is unexplored.

\section{Method}
\label{sec:method}

\subsection{Task Definition and Overview}
\label{sec:definition}

In this paper, we define the LS task, which combines the procedure of CWI and SG, as follows:
The goal of LS is to build a system that can convert a complex  text $X_i$ labeled by $s_x$ to a simple text $Y_i$ labeled by $s_y$ via replacing certain words with simpler words.
Of note, we leverage non-parallel corpora, which is denoted as $\mathcal{D}_x =\{X_i\}$ and $\mathcal{D}_y = \{Y_i\}$, to build an LS system, thereby bypassing parallel corpora.

\begin{figure*}[!ht]
\centering
\includegraphics[width=0.99\textwidth]{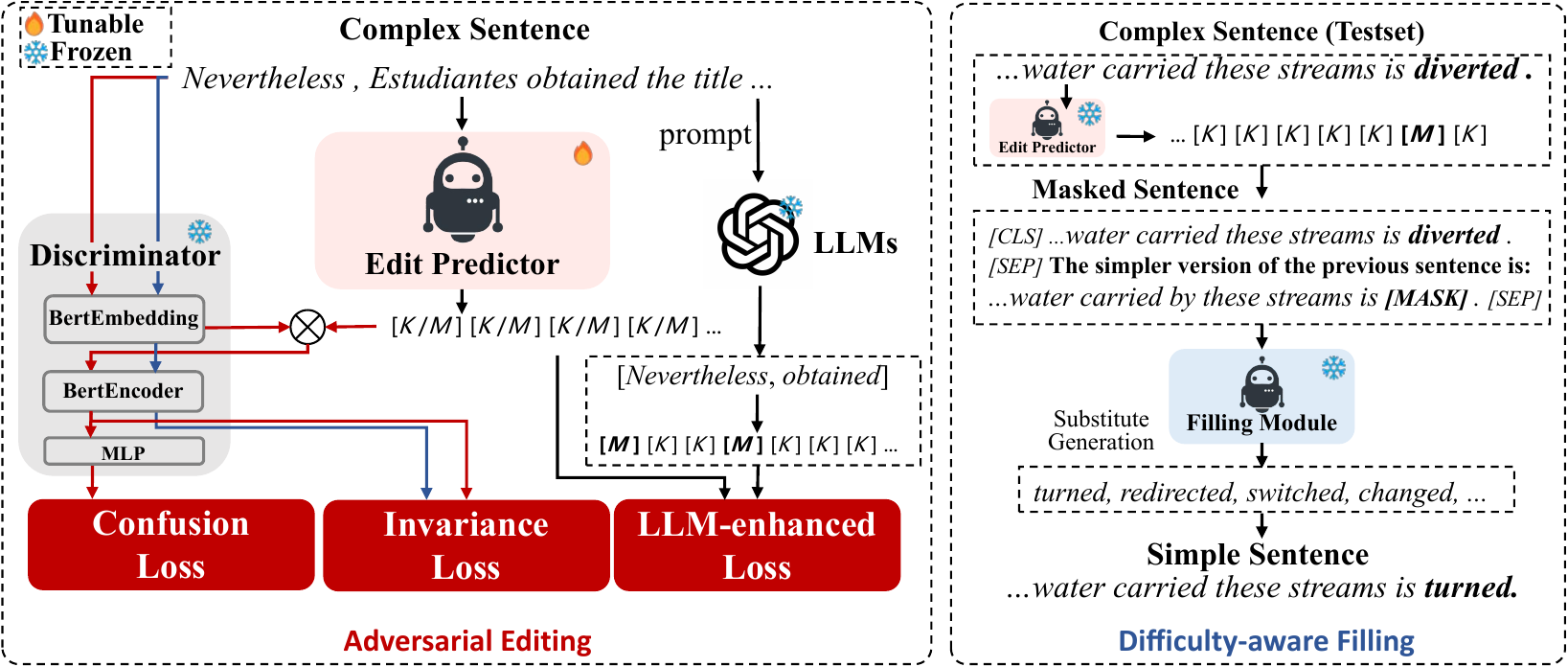} 
\caption{
The overall architecture of \textbf{LAE-LS}. \textit{Left panel}: \textbf{Adversarial Editing} module, where an Edit Predictor predicts lexical edits guided by confusion loss, invariance loss, and LLM-enhanced loss. \textit{Right panel}: \textbf{Difficulty-aware Filling} module, in which a Filling Module combines complex sentences and lexical edits into masked sentences as well as generates substitutions for the masked positions, aiming to simplify sentence.
}
\label{fig:architecture}
\end{figure*}

Our proposed method LAE-LS, which performs lexical simplification, involves two modules: \textbf{Adversarial Editing} and \textbf{Difficulty-aware Filling}. The overall architecture of our method is displayed in Fig.~\ref{fig:architecture}. 
Specifically, the Adversarial Editing module is an edit-based generative adversarial network designed to make lexical edits, which can be used to mask the complex words in complex sentences.
Built upon the well-trained Adversarial Editing module, a Difficulty-aware Filling module is used to fill in the masked position with simple words.
Remarkably, unlike the previous filling model \cite{qiang2021lsbert}, the Difficulty-aware Filling module, which is a cloze model, not only considers original sentences as clues but also maintains an awareness of producing simpler words.

\subsection{Adversarial Editing}
In this subsection, we detail the Adversarial Editing module.
Compared to existing efforts~\citep{laban2021keep,zhao2020semi,surya-etal-2019-unsupervised}, which directly model the text simplification from $X_i$ to $Y_i$, our module formulates the lexical simplification as an editing task, the goal of which is to predict lexical edits.
As shown in Fig.~\ref{fig:architecture}, we aim to train an Edit Predictor to perform the said procedure.

\subsubsection{Edit Predictor and Discriminator}
Before proceeding formally, we give notations for descriptive convenience as well as present Edit Predictor and Discriminator. 
We add a special token ``[\textit{CLS}]'' at the beginning of the sentence and denote a sentence by $X_i = [t_{i,1}, t_{i,2}, ..., t_{i,L}]$, where $t_{i,l}$ ($l\in[L]$) is $l$-th token in $X_i$ and $L$ is the length of the sentence.
For Edit Predictor, given that $X_i$, its output is $G_i = [g_{i,1}, g_{i,2}, ..., g_{i,L}]$, where $g_{i,l} \in \{\textit{K}, \textit{M}\}$~($l\in[L]$) is the edit operation for $t_{i,l}$ in $X_i$.
Note that ``\textit{K}'' indicates the current token is not a stylistic token or is already a simple token, and ``\textit{M}'' indicates the current token should be replaced by a mask token.
For example, the edit operation for a complex sentence ``[\textit{Much, of, the, water, carried, by, these, streams, is, diverted, .}]'' is ``[\textit{K}, \textit{K}, \textit{K}, \textit{K}, \textit{K}, \textit{K}, \textit{K}, \textit{K}, \textit{K}, \textit{M}, \textit{K}]''.
In our experiments, the Edit Predictor is built upon a  BERT~\cite{devlin2019bert} encoder.
To be specific,
given that $X_i$, we encode them with a sequence of hidden representations and then predict the edit labels via a Multilayer Perceptron (MLP) as follows:
\begin{align}
    & \mathbf{w}_l = \mathbf{w}_l^{tok} + \mathbf{w}_l^{typ} + \mathbf{w}_l^{pos} (l\in [L]), \nonumber \\
    & \mathbf{H} = [\mathbf{h}_1, 
    \mathbf{h}_2, \cdots, \mathbf{h}_L] = \text{BERT}(\mathbf{w}_1, 
    \mathbf{w}_2, ..., \mathbf{w}_L), \nonumber\\
    & P_G =  \text{softmax} (\mathbf{W}\mathbf{H} + \mathbf{A}), \label{eq:p_g} 
\end{align}
where $\mathbf{w}_l^{tok}$, $\mathbf{w}_l^{typ}$, $\mathbf{w}_l^{pos}$ represent the word embedding, token type embedding, and position embedding of $l$-th token, respectively. 
$\mathbf{W}$ and $\mathbf{A}$ are learnable parameters,
$P_G $ is a sequence of probabilities for a sentence.
Each element represents the ``\textit{K}/\textit{M}'' probability for a certain token.

For the discriminator, we still leverage BERT to encode it and use the first token in the last layer to represent the sentence, i.e., $\mathbf{h}_1$.
A fully connected layer is employed to help the model identify the style of the sentence:
\begin{align}
    P_D = \text{Classifier}(\mathbf{h}_1) = \text{softmax}(\mathbf{V}\mathbf{h}_1 + \mathbf{b}),
\end{align}
where $\mathbf{V}$ and $\mathbf{b}$ are learnable parameters, $P_D$ is the predicted probability of style $\{s_x, s_y\}$.
As the traditional adversarial network, the objective of the discriminator is to predict the style of the sentences in corpora correctly:
\begin{align}
\mathcal{L}_{D} =-\mathbb{E_{ \mathit{D, u\sim \left \{\mathcal{D}_x, \mathcal{D}_y \right \}} }} [\log_{P_{D}}(s_u|u)], 
\end{align}
where $u$ is the sentence sampled from $\mathcal{D}_x$ or $\mathcal{D}_y$ and $s_u$ is the true label for $u$.
Note that the discriminator needs to be trained in advance, and will be frozen in subsequent training.

\subsubsection{Training}
\label{sec:training}

Regarding traditional training \citep{surya-etal-2019-unsupervised} for adversarial generation, given a complex sentence $X_i$, the output of the generator is the simplified version, denoted as $\hat{Y}_i$. 
In this case, the discriminator tries to distinguish whether the sentence is simple or not, and the generative network is trained to fool the discriminator, making it difficult to differentiate between simple and complex. 
However, the above loss is not applicable in our framework due to the following two issues: 
\begin{enumerate}
\item It is not feasible to include the raw output of the Edit Predictor for adversarial training as ``\textit{K}'' and ``\textit{M}'' cannot be directly encoded by the discriminator.
\item It is vital to control the predicted edits and maintain the syntax for lexical simplification. 
However, existing methods usually ignore this and lead to unexpected changes to the original sentences.
\end{enumerate}

To solve the above issues, we elaborate the objective of Edit Predictor with respect to \textit{confusion loss}, \textit{invariance loss}, and \textit{LLM-enhanced loss}. In other words, the training of the Edit Predictor is guided with the consideration of measuring the confusion degree to the well-performing pre-trained discriminator (\textit{confusion loss}), semantic invariance to the original sentence (\textit{invariance loss}) and similarity to the LLMs' signals (\textit{LLM-enhanced loss}), respectively.

\noindent \textbf{Confusion Loss.} To commence, we investigate guiding Edit Predictor in editing complex tokens within a sentence. 
We aspire to integrate the output of Edit Predictor into the discriminator.
To achieve it, 
we can characterize the sentence representation with the edit labels.
Specifically, we multiply the token embedding with the probability of ``\textit{K}'' predicted by the Edit Predictor, and assume a sentence without a stylistic token will confuse the discriminator by producing a moderate judgement of the transferred sentence.

Formally, the said operation can be represented as: 
\begin{align}
    & \mathbf{w}_l = \mathbf{w}_l^{tok}\cdot p_l^{K} + \mathbf{w}_l^{typ} + \mathbf{w}_l^{pos}(l\in[L]), \notag \\
    & \mathbf{H}^{conf}= \text{BERT}(\mathbf{w}_1, \mathbf{w}_2, ..., \mathbf{w}_L), \notag \\
    & P_D = \text{Classifier}(\mathbf{h}_1^{conf}),
\end{align}
where $p_l^{K}$ is the predicted probability of ``\textit{K}'' derived from Equation~(\ref{eq:p_g}), and $\mathbf{h}_1^{conf}$ is the first token embedding.
If a token has a low probability of ``\textit{K}'', the semantic information of the token will be removed from the sentence.
For this purpose, we frame a confusion loss $\mathcal{L}_G^{conf}$ is defined below:
\begin{align}
    \mathcal{L_{\mathit{G}}^{\mathit{conf}}} = (P_{D} - \alpha)^2.
\end{align}

Here, $\alpha$ is a hyper-parameter that we set as an unconfident score.
We assume a masked sentence would confuse the discriminator by showing an unconfident score around $\alpha$ instead of \{$0$, $1$\}.

\noindent \textbf{Invariance Loss.} 
Regarding the second issue, we incorporate an invariance loss.
This additional loss function serves as a reference, encouraging the Edit Predictor to focus on preserving the underlying semantic information of the sentence.
The invariance loss, denoted as $\mathcal{L_{\mathit{G}}^{\mathit{inv}}}$, is computed using the Cosine Similarity metric as outlined below.
\begin{align}
\mathcal{L_{\mathit{G}}^{\mathit{inv}}} = 1-cos(\mathbf{h}_1, \mathbf{h}^{conf}_1).
\end{align}

As we can see, when the semantics of the sentences are preserved, the loss tends towards $0$, otherwise, it approaches $1$.

\noindent \textbf{LLM-enhanced Loss}.
Large Language Models (LLMs)~\cite{brown2020language}, such as ChatGPT or OPT~\cite{zhang2022opt}, have proven highly effective. 
A series of approaches manage to leverage LLMs to provide external knowledge for various NLP tasks, such as Question Answering and Information Retrieval~\cite{lan2023improving,guo2023images,kojima2022large,wei2022chain,trivedi_acl2023}.
As shown in Fig.~\ref{fig:architecture}, given the sentence ``\textit{Nevertheless , Estudiantes obtained the title at the end of the Apertura 2006 .}'', LLMs help to annotate complex words ``\textit{Nevertheless}'' and ``\textit{obtained}'' serving as signals for the Edit Predictor. 
Motivated by this, we introduce LLM knowledge into the Adversarial Editing System by prompting LLMs, extracting the supervision signals from the responses, and fusing them into the adversarial training procedure.

However, as mentioned above, LLMs are likely to make unexpected edits to the syntax of the sentences.
To circumvent over-edit to the original sentences, instead of rewriting the sentences, we prompt LLMs with the instruction as follows:

\begin{small}
\vspace{0.2cm}
\hrule
\vspace{0.1cm}
\begin{Verbatim}[commandchars=\\\{\}, breaklines=true]
\textbf{Prompt:}
\hspace{1em} Please identify the complex words in the following sentence.
\textbf{Sentence:} 
\hspace{1em} \{Input Sentence\}
\textbf{Output format:} 
\hspace{1em} [w1, w2, ...]
\end{Verbatim}
\hrule
\vspace{0.2cm}
\end{small}

\noindent In light of the above instruction, LLMs are required to produce complex tokens. 
Thus, LLM pseudo label loss (termed LLM-enhanced Loss) takes the form:
\begin{align}
\mathcal{L_{\mathit{G}}^{\mathit{LLM}}} = - \frac{1}{L}\sum_{w_l \in X_{i}} [\log_{P_G}(g^*_l | w_l)],
\end{align}
where $g^*_l$ is the pseudo gold edit of token $w_l$ generated by LLMs.

Eventually, the overall loss function for the edit predictor is:
\begin{align}
\mathcal{L_{\mathit{G}}} = \lambda _1\mathcal{L_{\mathit{G}}^{\mathit{conf}}}+\lambda _2\mathcal{L_{\mathit{G}}^{\mathit{inv}}}+\lambda _3\mathcal{L_{\mathit{G}}^{\mathit{LLM}}},
\end{align}
where $\lambda _1$, $\lambda _2$ and $\lambda _3$ are tunable hyperparameters for balancing different loss items.
There is an advantage of distilling the knowledge from LLMs to the Edit Predictor instead of directly leveraging LLMs to make the prediction.
That is, LLMs have the risk of over-editing, taking their outputs as the supervision signals play the effect of distilling high-quality knowledge to the small-size models, which can effectively restrain the over-fitting issue~\cite{gou2021knowledge}.

\subsection{Difficulty-Aware Filling}
On top of the well-trained Edit Predictor, which generates a sequence of edit operations, we keep the tokens with ``\textit{K}'' unchanged and mask the tokens with ``\textit{M}''. 
In the example shown in Fig.~\ref{fig:architecture}, 
we obtain ``[\textit{Much, of, the, water, carried, by, these, streams, is, M}]'' via Edit Predictor.
We denote it as $\tilde{X}_i$.

The next procedure is to replace the ``\textit{M}'' labels with simplified tokens.
Existing studies leverage pretrained models to fill in the slots based on the context~\citep{wang2022text, sun2023teaching}. 
However, for text simplification, merely including the masked sentences $\tilde{X_i}$ makes the pretrained models ignore the semantic meaning of the complex tokens in the original sentences.
Recent effort ~\cite{qiang2021lsbert} inputs both $X_i$ and $\tilde{X}_i$ as a pair and feed them into the pre-trained models to predict the tokens at the ``\textit{M}'' positions in $\tilde{X}_i$.
This facilitates pre-trained models to generate tokens with the same semantic meaning as the original complex tokens.

Inspired by this, we introduce a Difficulty-aware Filling module by placing the prompt ``\textit{The simpler version of the previous sentence is: }'' in between $X_i$ and $\tilde{X}_i$ to encourage the pre-trained model to be aware of the change of the difficulty level. 
The prompt for the Difficulty-aware Filling module is shown below.

\begin{small}
\vspace{0.2cm}
\hrule
\vspace{0.1cm}
\begin{Verbatim}[commandchars=\\\{\}, breaklines=true]
\textbf{Difficulty-aware Filling Prompt: }
\hspace{1em} [CLS] Original sentence [SEP] \textbf{The simpler version of the previous sentence is: } Masked sentence [SEP]
\rule{\linewidth}{0.2pt}
\textbf{Example: }
\hspace{1em} [CLS] much of the water carried these streams is diverted . [SEP] \textbf{The simpler version of the previous sentence is:} much of the water carried by these streams is \textbf{\scriptsize{[MASK]}} . [SEP]
\end{Verbatim}
\hrule
\vspace{0.2cm}
\end{small}

By virtue of the mentioned procedure, we feed the input into the BERT model and let the model predict the words at mask positions.
Intuitively, we denote the module as:
\begin{align}
    \hat{Y}_i = \text{Filling\_Module}(\tilde{X}_i).
\end{align}

We extract the predicted sentence followed by the instruction as the final simplified sentence $\hat{Y}_i$.

\section{Experiments}
\label{sec:exp}

\subsection{Experimental Settings}
\noindent \textbf{Datasets.}
To gauge the effectiveness of our proposed approach, we employ three commonly used datasets for the LS task, namely \textbf{LexMTurk} \citep{horn2014learning}, \textbf{BenchLS} \citep{paetzold2016benchmarking}, and \textbf{NNSeval} \citep{paetzold2016unsupervised}.
These three datasets contain 500, 929, and 239 testing samples, respectively.
Also, each sentence in the datasets is annotated with complex words, and multiple alternative simplified words are provided.
It is noteworthy that we work on the LS task without parallel corpora thus tap \textbf{WikiSmall}~\citep{zhu2010monolingual} as the non-parallel corpus, i.e., $\mathcal{D}_x$ and $\mathcal{D}_y$, which encompasses $84,296$ complex sentences and $84,296$ simple sentences, respectively.

\noindent \textbf{Baselines.}
We compare our method with a wide range of baselines for CWI, SG, and LS tasks.

1) Evaluation on CWI:
Following previous work~\citep{yimam-etal-2017-cwig3g2}, we use \textbf{Character}, \textbf{Syllable}, \textbf{Vowel}, \textbf{Frequency} features and identify complex words via setting a threshold. 
Moreover, \textbf{Attention}~\cite{wang2022text} from a well-trained discriminator can be utilized as it usually assigns more scores to the complex tokens.
Additionally, \textbf{LSBert} \cite{qiang2021lsbert} employs BERT as a sequence labeling model and undergoes supervised training on the CWI 2018 dataset \citep{yimam2018report}.

2) Evaluation on SG:
A line of methods is given complex words and simply predicts the simplified words.
Early methods like \textbf{Paetzold-CA} \citep{paetzold2016unsupervised}, \textbf{Paetzold-NE} \citep{paetzold2017lexical}, and \textbf{REC-LS} \citep{gooding2019recursive} utilize word embeddings and rely heavily on parallel corpora or WordNet for assistance.
Recently, \textbf{LSBert} \citep{qiang2021lsbert}, \textbf{BART} \cite{lewis2020bart} and \textbf{SimpleBART} \citep{sun2023teaching} leverage the mask word prediction capability of pretrained models.

3) Evaluation on LS:
Due to the lack of LS methods that integrate CWI and SG, we design several baselines (e.g., \textbf{Character-LSBert}, \textbf{Syllable-LSBert}, \textbf{Vowel-LSBert}, \textbf{Frequency-LSBert}, \textbf{Attention-LSBert}) for the two stages, building upon \textbf{LSBert}~\cite{qiang2021lsbert}.

\noindent \textbf{Evaluation Metrics.}
To fairly compare the performance of different methods, we follow the standard evaluation metrics to measure the CWI and SG~\citep{qiang2021lsbert}.
Also, we introduce an evaluation metric to evaluate the performance of varying LS systems, which takes two-stage accuracy into consideration.
Concretely, we identify a correct prediction when both the complex and top-ranked simplified words are predicted correctly.
Hence, we calculate the Precision, Recall and F1 for each sentence and compute the average scores for each test set.

\noindent \textbf{Implementation Details.}
Unless otherwise stated, we set $\lambda_1$, $\lambda_2$ and $\lambda_3$ all to $1$. We fix the unconfident score $\alpha = 0.5$ as default.
For Discriminator, we train it with the whole WikiSmall dataset and yield a well-performing Discriminator, which achieves an accuracy of $96.44\%$ on the development set and will be frozen in subsequent training.
For Edit Predictor, only complex sentences from the WikiSmall dataset are leveraged.
For the Difficulty-aware Filling module, we remove non-English characters and the morphological derivations of the complex words and choose the top $10$ words as the substitution following\citep{qiang2021lsbert}.
All the models are trained for $30$ epochs with batch size of $32$. 
We use Adam optimizer \citep{kingma2014adam} with learning rate of $1e$-$5$.
Similar with the setting of prior work~\citep{omelianchuk2021text}, we freeze the BERT layer weights during the first four epochs of training, and perform early stopping after $3$ epochs account for the performance on the development set.
We obtain the SG results of Paetzold-CA, Paetzold-NE, REC-LS, LSBert from \cite{qiang2021lsbert} and BART, SimpleBART from \cite{sun2023teaching}, while the remaining models are re-implemented by us.

\subsection{Experiment Results}

\begin{table*}[t!]
    \centering
    \small
\begin{tabular}{l | c c c | c c c | c c c}
        \toprule  
        \multirow{2}{*}{\centering\textbf{Methods}} & \multicolumn{3}{c|}{LexMTurk} & \multicolumn{3}{c|}{BenchLS}  & \multicolumn{3}{c}{NNSeval}\\
        \cmidrule(lr){2-4} \cmidrule(lr){5-7} \cmidrule(lr){8-10}
        & Precision & Recall & F1 & Precision & Recall & F1 & Precision & Recall & F1 \\
        \midrule
        \multicolumn{10}{c}{\centering\textit{Complex Word Identification}} \\
        \midrule
        Character &  0.122 &  0.780 & 0.211 & 0.111 & 0.755 & 0.194 & 0.105 & 0.716 & 0.183 \\
        
        Syllable & 0.140 &  0.606 & 0.228 & 0.117 & 0.526 & 0.191 & 0.100 & 0.456 & 0.163 \\

        Vowel & 0.132 &  0.764 & 0.226 & 0.117 & 0.727 & 0.201 & 0.108 & 0.678 & 0.186 \\

        Frequency & 0.078 &  0.632 & 0.139 & 0.072 & 0.623 & 0.129 & 0.054 & 0.456 & 0.096 \\
        
        Attention & 0.064 & 0.512 & 0.114 & 0.062 & 0.448 & 0.109 & 0.058 & 0.435 & 0.103 \\

        LSBert & \textbf{0.136} & 0.795 & 0.231 & \textbf{0.136} & 0.788 & \textbf{0.231} & 0.121 & 0.707 & 0.207 \\
        
        LAE-LS (ours) & 0.135 & \textbf{0.810} & \textbf{0.232} & 0.128 & \textbf{0.813} & 0.221 & \textbf{0.126} & \textbf{0.824} & \textbf{0.218} \\
        \midrule
        \multicolumn{10}{c}{\centering\textit{Substitute Generation}} \\
        \midrule
        Paetzold-CA & 0.177 & 0.140 & 0.156 & 0.180 & 0.252 & 0.210 & 0.118 & 0.161 & 0.136 \\
        Paetzold-NE & 0.310 & 0.142 & 0.195 & 0.270 & 0.209 & 0.236 & 0.186 & 0.136 & 0.157 \\
        REC-LS & 0.151 & 0.154 & 0.152 & 0.129 & 0.246 & 0.170 & 0.103 & 0.155 & 0.124 \\
        LSBert & 0.306 & 0.238 & 0.268 & 0.244 & 0.331 & 0.281 & 0.194 & 0.260 & 0.222 \\
        BART & 0.192 & 0.183 & 0.188 & 0.196 & 0.178 & 0.192 & - & - & - \\
        SimpleBART & 0.287 & \textbf{0.282} & 0.285 & \textbf{0.280} & 0.276 & 0.278 & - & - & - \\
        LAE-LS (ours) & \textbf{0.340} & 0.264 & \textbf{0.297} & 0.262 & \textbf{0.355} & \textbf{0.301}& \textbf{0.202} & \textbf{0.269} & \textbf{0.231} \\
        \midrule
        \multicolumn{10}{c}{\centering\textit{Lexical Simplification}} \\
        \midrule

        Character-LSBert & 0.080 &  0.540 & 0.139 & 0.061 & 0.434 & 0.107 & 0.044 & 0.318 & 0.078 \\

        Syllable-LSBert & 0.090 &  0.410 & 0.148 & 0.064 & 0.299 & 0.105 & 0.042 & 0.201 & 0.069\\

        Vowel-LSBert & 0.087 &  0.528 & 0.149 & 0.063 & 0.412 & 0.110 & 0.045 & 0.293 & 0.077\\

        Frequency-LSBert & 0.047 &  0.440 & 0.085 & 0.036 & 0.364 & 0.066 & 0.023 & 0.226 & 0.042\\
        
        Attention-LSBert & 0.039 &  0.350 & 0.070 & 0.031 & 0.243 & 0.054 & 0.020 & 0.167 & 0.036\\

        LSBert & \textbf{0.097} &  0.564 & 0.166 & 0.075 & 0.454 & 0.129 & 0.056 & 0.335 & 0.095\\

        LAE-LS (ours) & \textbf{0.097} &  \textbf{0.582} & \textbf{0.167} & \textbf{0.077} & \textbf{0.489} & \textbf{0.133} & \textbf{0.058} & \textbf{0.381} & \textbf{0.101}\\
        
        \bottomrule
    \end{tabular}
    \caption{CWI, SG and LS evaluations on three benchmark datasets. 
    }
    \label{table: Main Results}
\end{table*}

\begin{table}[htb]
    \centering
    \small
    \begin{tabular}{l | c c c c}
        \toprule
        & Size & F1-CWI & F1-SG & F1-LS \\
        \midrule

        {\small ChatGLM2} & $6$B & 0.027 & 0.250 & 0.048  \\

        {\small llama2} & $13$B & 0.115 & 0.264 & 0.085  \\
        
        {\scriptsize GPT-3.5-turbo} & $175$B & 0.221 & 0.296 & \textbf{0.200} \\
        
        {\small LAE-LS (ours)} & $220$M & \textbf{0.232} & \textbf{0.297} & 0.167\\
        \bottomrule

    \end{tabular}
    \caption{Comparison with various LLMs on LexMTurk Datasets in term of parameter size and F1.
    }
    \label{table: Comparison with LLMs}
\end{table}

\begin{table}[t!]
    \centering
    \small
\begin{tabular}{l | c c c }
        \toprule  
        & F1-CWI & F1-SG & F1-LS \\
        \midrule
        LAE-LS (baseline)  & \textbf{0.232} & \textbf{0.297} & \textbf{0.167} \\
        \midrule
        
        {\scriptsize w/o LLM-enhanced Loss}  & 0.094 & 0.297 & 0.066\\

        {\scriptsize w/o Confusion Loss} & 0.078 & 0.297  & 0.135 \\

        {\scriptsize w/o Invariance Loss} & 0.089 & 0.297 & 0.153\\
        
        {\scriptsize w/o Difficulty-aware Filling} & 0.232 & 0.268  & 0.162 \\
        
        \bottomrule
    \end{tabular}
    \caption{Ablation Study of LS on LexMTurk Datasets w.r.t. F1.
    }
    \label{table: Ablation Study}
\end{table}

\begin{table*}[!t]
    \centering
    \small
    \begin{tabular}{l | l }
        \toprule 
        \textbf{Methods} & Sentence \\
        \midrule
        Sent (1) & Triangles ... be \textbf{classified} according to their internal angles, measured here in degrees. \\
        Candidates &  \{called, labeled, divided, coded, defined, listed, \textbf{categorized}, named, organized, described...\}\\
        \midrule
        LSBert & \textbf{squares} ... be \textbf{categorized} according to their \textbf{external} \textbf{triangles}, \textbf{} here in \textbf{metric}. \\
        {\scriptsize GPT-3.5-turbo} & Triangles ... be \textbf{categorized} according to their \textbf{inner} \textbf{corners}, \textbf{calculated} here in \textbf{units}. \\
        LAE-LS (ours) & Triangles ... be \textbf{categorized} according to their internal angles, measured here in degrees. \\

        \midrule
        \midrule
        Sent (2) & Stone floor tiles tend to ... ceramic tiles and somewhat more \textbf{prone} to breakage... \\
        Candidates & \{liable, easier, probable, subject, \textbf{susceptible}, disposed, \textbf{likely}, inclined, vulnerable, apt...\} \\
        \midrule
        LSBert & Stone floor tiles tend to ... \textbf{porcelain} tiles and somewhat more \textbf{susceptible} to \textbf{cracking}... \\
        {\scriptsize GPT-3.5-turbo} & Stone floor tiles tend to ... \textbf{clay} tiles and somewhat more prone to \textbf{damage}... \\
        LAE-LS (ours) & Stone floor tiles tend to ... ceramic tiles and somewhat more \textbf{likely} to breakage... \\
        \bottomrule
    \end{tabular}
    \caption{Case study of LS on LexMTurk Datasets.
    Complex words are highlighted in bold.
    Candidates indicate the list of annotated simple words for the corresponding complex words in the dataset.
    Differences between generated and original sentences are in bold.
    }
    \label{table: Case study}
\end{table*}

\subsubsection{Results Comparison}
\textbf{Comparison with Baselines.}
We study the performance of different methods for CWI, SG and LS tasks on LexMTurk, BenchLS and NNSeval datasets, as shown in Table~\ref{table: Main Results}. 
The results show that:
(1) For the CWI task, which assesses the abilities of models to identify complex words within sentences, LAE-LS achieves the best results on the LexMTurk and NNSeval datasets and demonstrates competitive performance on the BenchLS dataset.
One can see that using features (e.g., word character length and syllables) as the simple factors for identifying complex words does not yield satisfactory results.
It's worth noting that LSBert is under supervised training, while LAE-LS surpasses it without any annotated parallel corpora.
(2) 
In the SG task, our method outperforms all baselines on the three datasets, thus validating the effectiveness of the Difficulty-aware Filling module.
Compared with LSBert, which solely takes a pair of original and masked sentences as the input, LAE-LS exhibits significantly superior performance.
This verifies that placing prompts between sentence pairs helps the model be aware of the difficulty for words.
(3) 
Regarding the LS task, our method consistently outperforms the baselines when we integrate CWI and SG together.
It is remarkable that there is a performance decrease in terms of F1 for the LS task, compared to CWI and SG tasks.
This is because we identify a correct prediction only when complex words are correctly identified and simplified simultaneously, thereby increasing the difficulty of the task.
Moreover, we observe that the precision is low for all methods since these methods typically predict more complex words, even though only one is labeled.
As such, how to augment the precision in the LS task is challenging. 

\noindent \textbf{Comparison with LLMs.}
Here, we explore the performance of our method and popular LLMs with more parameter size in terms of F1 over LexMTurk dataset.
In this end, we select three LLMs with different parameter sizes, including \texttt{ChatGLM2}~\cite{du2021glm}\footnote{https://huggingface.co/THUDM/chatglm2-6b}, \texttt{llama2}~\cite{touvron2023llama}\footnote{https://huggingface.co/meta-llama/Llama-2-13b-chat-hf}, \texttt{GPT-3.5-turbo}~\cite{brown2020language}\footnote{https://openai.com/}.
Note that we either employ API or download the checkpoints to do the prediction, and take the responses from the LLMs as the prediction.

Specifically, we construct the following prompts for the CWI task:

\begin{small}
\vspace{0.2cm}
\hrule
\vspace{0.1cm}
\begin{Verbatim}[commandchars=\\\{\}, breaklines=true]
\hspace{1em}  Please identify the complex words in the following sentence. Sentence: \{Input Sentence\}.
\end{Verbatim}
\hrule
\vspace{0.2cm}
\end{small}
Similarly, we construct the following prompts for the SG task:
\begin{small}
\vspace{0.2cm}
\hrule
\vspace{0.1cm}
\begin{Verbatim}[commandchars=\\\{\}, breaklines=true]
\hspace{1em}  Please provide 10 simplified alternative words for the word \{Complex Word\} in the sentence. Sentence: \{Input Sentence\}.
\end{Verbatim}
\hrule
\vspace{0.2cm}
\end{small}
For the LS task, we regard concatenating the results from the two sub-tasks (including CWI and SG) as it results.
The results are displayed in Table~\ref{table: Comparison with LLMs}.

As we can see from the table, LAE-LS, which has a smaller parameter size, can achieve competitive results comparing with the powerful LLMs.
For both CWI and SG tasks, LAE-LS trumps all LLMs in terms of F1, indicating that its domination in identifying complex words within sentences and predicting substitutes for each complex word.
For the LS task, with respect to F1, LAE-LS leads \texttt{ChatGLM2} and \texttt{llama2}, but falls slightly behind \texttt{GPT-3.5-turbo}.
This is because when we compute the F1 score of LS tasks, only the top ranked substitute is taken into consideration while top $10$ substitutes are used to measure the F1 score for SG tasks.
This indicates that \texttt{GPT-3.5-turbo} has the advantage of generating the accurate substitute with top $1$ prediction.
To sum up, LAE-LS manages to achieve competitive results with a considerably smaller number of parameters compared to \texttt{GPT-3.5-turbo}.

\subsubsection{Ablation Study}
In this section, we systematically delve into the effect of the components in our method by performing the leave-one-out test. 
Specifically, we iteratively remove the loss functions defined 
in our Adversarial Editing module. 
Also, we look into the effectiveness of the Difficulty-aware Filling module.
The results are shown in Table~\ref{table: Ablation Study}.
From the experimental results, it is evident that removing any of these loss functions leads to performance drop, suggesting that they are vital for the training of Edit Predictor.
As we have introduced in Section~\ref{sec:training}, they measure the prediction from different aspects.
Of note, LLM-enhanced loss plays the most crucial role for training of Edit Predictor.
This means that it is effective to distill the knowledge from LLMs to models with smaller size, especially in the absence of annotated data.
Moreover, we replace the Difficulty-aware Filling module with LSBert and the results show there is a notable decrease on the F1 score of the SG task.
The above results indicate that our proposed Difficulty-aware Filling module indeed guides the pretrained model to generate accurate words rather than words that do not preserve the original meaning or not simple enough.

\subsubsection{Case Study}
To further go into the virtue of LAE-LS, we present multiple case studies in Table~\ref{table: Case study}.
For sentence (1), we can observe that all models successfully identify ``\textit{classified}'' as a complex word and replace it correctly with ``\textit{categorized}''.
However, in the case of LSBert, it identifies ``\textit{triangles}'' as a complex word, which is inaccurate because ``\textit{triangles}'' is a non-stylistic word that holds the semantic information of the sentence. 
We suspect that LSBert is trained on supervised CWI data, focusing on the inherent difficulty of words and ignoring the semantic meaning of them in the sentence. 
In contrast, our Edit Predictor is measured by the invariance loss, which preserves the semantic information of the original sentence effectively.
Notably, compared with \texttt{GPT-3.5-turbo}, which tends to identify more complex words and over-simplify the sentence, LAE-LS incorporates adversarial training, which enables to more accurate complex words identification and prevents excessive modifications in the sentence.
For sentence (2), apart from \texttt{GPT-3.5-turbo}, both LAE-LS and LSBert accurately identify this complex word, though LSBert tends to label many other words as complex words.
Significantly, LSBert simplifies \textit{"prone"} to "susceptible," while LAE-LS simplifies it with a much simpler word \textit{``likely''}.
This suggests that our method considers not only the semantic meaning of the context but also the complexity of the generated words, thus leading to a more desirable prediction.

\section{Conclusion}
\label{sec:cons}
In this paper, we propose an LLM-enhanced Adversarial Editing System to address the lexical simplification task without parallel corpora, which consists of an Adversarial Editing module and a Difficulty-aware Filling module.
Adversarial Editing module is guided by a confusion loss and an invariance loss to make lexical edits with a consideration of semantic preservation and simplified ratio. 
Meanwhile, we craft an LLM-enhanced loss to distill knowledge from LLMs, thus further augmenting the Adversarial Editing module.
From that, the Difficulty-aware Filling module combines the original sentences and lexical edits to mask complex words within sentences and fill in the masked positions with simpler words.
The extensive experimental results on three LS datasets demonstrate that our method is effective.
That is, our method not only advances lexical simplification in the absence of parallel corpora but also showcases the potential for leveraging the capabilities of large language models to enhance the simplification process.

\section*{Ethics Statement}
This is a study about Lexical Simplification.
It does not have any data privacy issues. 
We did not collect any personal information. 
This is a task that involved no risk as the participants were not exposed to any harmful material or asked to perform any risky tasks. 

\section*{Acknowledgements}
This work was supported by Joint Key Project (Project No. U23A20298) and Young Scientists Project (Project No. 62206097) of National Natural Science Foundation of China, Shanghai Pujiang Talent Program (Project No. 22PJ1403000) and East China Normal University (Project No. 2022ECNU—WHCCYJ-31).

\section*{Bibliographical References}\label{sec:reference}
\bibliographystyle{lrec-coling2024-natbib}
\bibliography{lrec-coling2024-example}

\end{document}